\pdfoutput=1
\documentclass[11pt]{article}

\usepackage[]{acl}

\usepackage{times}
\usepackage{latexsym}

\usepackage[T1]{fontenc}
\usepackage[utf8]{inputenc}

\usepackage{microtype}

\usepackage{inconsolata}

\usepackage{natbib}
\usepackage{csquotes}
\usepackage{graphicx}
\usepackage{amsmath}
\usepackage{amsthm}
\usepackage{amssymb}
\usepackage{algorithm}
\usepackage{algorithmic}
\usepackage{multirow}
\usepackage{colortbl}
\definecolor{mygray}{gray}{.9}
\usepackage{booktabs}
\usepackage{bbm}
\usepackage{subfigure}
\usepackage{hyperref}
\usepackage{array}
\usepackage{makecell}
\newcommand{\C}{\boldsymbol{C}}
\newcommand{\Q}{\boldsymbol{Q}}
\newcommand{\Po}{\boldsymbol{P}}
\newcommand{\Ho}{\boldsymbol{H}}
\title{Concept Based Continuous Prompts for Interpretable Text Classification}

\author{Qian Chen$^1$, Dongyang Li$^1$, Xiaofeng He$^{1,2}$ $\thanks{\, Corresponding author}$
 \\
  $^1$ School of Computer Science and Technology, East China Normal University, Shanghai, China 
  \\ 
  $^2$ NPPA Key Laboratory of Publishing Integration Development, ECNUP, Shanghai, China \\
  \texttt{\{qianchen901005,fromdongyang\}@gmail.com,}  \texttt{hexf@cs.ecnu.edu.cn} \\}

\begin{document}
\maketitle
\begin{abstract}
Continuous prompts have become widely adopted for augmenting performance across a wide range of natural language tasks. However, the underlying mechanism of this enhancement remains obscure. 
Previous studies rely on individual words for interpreting continuous prompts, which lacks comprehensive semantic understanding.  
Drawing inspiration from Concept Bottleneck Models, we propose a framework for interpreting continuous prompts by decomposing them into human-readable concepts. 
Specifically, to ensure the feasibility of the decomposition, we demonstrate that a corresponding concept embedding matrix and a coefficient matrix can always be found to replace the prompt embedding matrix.
Then, we employ GPT-4o to generate a concept pool and choose potential candidate concepts that are discriminative and representative using a novel submodular optimization algorithm. 
Experiments demonstrate that our framework can achieve similar results as the original P-tuning and word-based approaches using only a few concepts while providing more plausible results.

\end{abstract}

\section{Introduction}
Continuous prompts are widely used in NLP  with  stable performance and easy-hand training compared to discrete prompts. However, the introduction of dense prompt tokens compromises the interpretability.

\begin{figure}[tbp]
    \centering
    \includegraphics[width=0.51\textwidth]{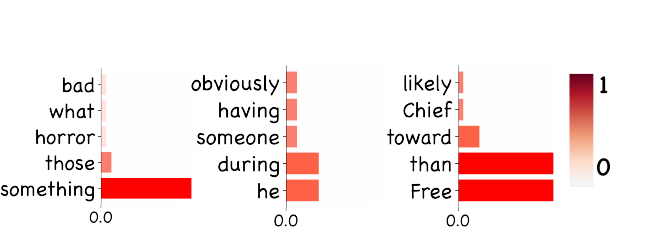}
    \caption{An explanation example from \citet{work3}. From left to right three sub-figures represent the  SST-2, IMDB,  and AGNews dataset results. The  color bar represents values.}
    \label{fig:intro1}
\end{figure}

Recent  trials  exploring the underlying mechanism found that models do not rely on context augmented by complementary prompt tokens, which is puzzling \cite{work0}. \citet{work1} suggest that  adding non-perceptible perturbation to prompt embedding could bring a  different outcome. \citet{work2} propose Prompt Waywardness Hypothesis, which means it is impossible to match a continuous prompt with one exact  discrete text. \citet{work3} introduce numerous words to mitigate the gap between discrete tokens and dense learned embedding. However, generated word explanations may lack comprehensible and integral semantics. For example, in Figure \ref{fig:intro1}, although given top-5 selected words, they are just listed without organic combination, being blended with stop words noise.

\begin{figure}[tbp]
    \centering
    \includegraphics[width=0.45\textwidth]{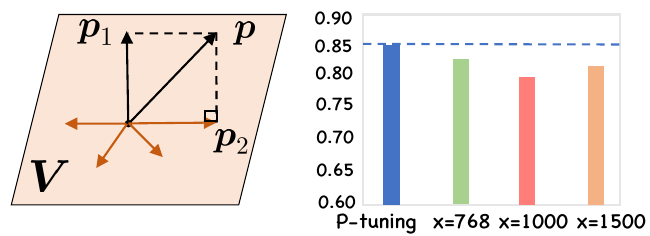}
    \caption{Left: The bad case from \citet{work3}, which the continuous prompt $\boldsymbol{p}$ fails to locate in the span space $\boldsymbol{V}$. Given the existence of $\boldsymbol{p}_{1}$, the fitting gap cannot be eliminated. Right: The accuracy results from \citet{work3}. x represents a vocabulary size. As the capacity of vocabulary increases,  the performance does not significantly change.}
    \label{fig:intro2}
\end{figure}

Moreover, due to the infeasibility of discrete words  regression to continuous prompt, the inherent gap would driven the abuse of words, as shown in Figure \ref{fig:intro2}, simply increasing vocabulary capacity fails to rise up the performance. And word-based methods cannot apply to BPE encoding \cite{bpe} based models directly like GPTs \cite{gpt-1,gpt-2}.

Inspired by Concept Bottleneck Models (CBMs) \cite{CBM}, we propose Concept Decomposition (CD) framework, which decomposes the prompt embeddings into  concepts. In detail, we start by reclaiming the formulation of the vanilla transformer block, in which we find that the outputs essentially correspond to linear combinations of column vectors from the outer layer of the Feed-Forward Network (FFN). Based on findings in \citet{transformers_key_value_memory}, we argue that  
continuous prompts could be decomposed into several comprehensible fragmented semantic patterns, which can be aligned with integral human-readable concepts. In order to complete the pattern-concept alignment, we first utilize the powerful GPT-4o \cite{gpt4} to construct an ample space of possible concepts owing to Large Language Models (LLMs) containing significant world knowledge that can be elicited by instructions instead of  labor-intensively annotation \cite{transformers_key_value_memory}. 

After construction, as we require highly discriminative concepts that cover various aspects to obtain a better explanation, we formulate  above requirements into a submodular optimization problem that allows us to choose good concepts efficiently \cite{submodular}. 

Extensive experiments demonstrate the effectiveness of our framework. In the evaluation of two typical model architectures (encoder-only and decoder-only) on three diverse datasets, CD achieves comparable performance to the original continuous prompt method both in few-shot and full-shot scenarios,  providing more meaningful and comprehensible concepts with superior to the discrete explanation method. 
% Meanwhile, we conduct experiments on closed domain settings to investigate the feasibility of efficiently enhancing knowledge with continuous prompts using our interpretations.

Our contributions are summarized as follows:
\begin{itemize}
    \item We prove that one can always find a concept embedding matrix and a coefficient matrix  corresponding to the continuous prompt matrix. 
    \item We propose a framework (CD) to maintain representative concepts  for interpreting continuous prompts.
    \item We evaluate the proposed framework on three classification datasets with two typical language models: BERT \cite{bert} and GPT-2 \cite{gpt-2}, and the results demonstrate the effectiveness.
\end{itemize}

\section{Related Work}

\subsection{Concept Bottleneck Models}
CBMs have garnered significant attention, with specific concepts serving as fundamental units of explanation, also known as rationales. These models solve for rationales and combine them to produce outputs. For instance, \citet{NEURIPS2020_ecb287ff} propose a concept discovery method that aims to infer a complete set of concepts. Recently, CBMs have been found to have a backdoor path due to inherent structures \cite{kumar-talukdar-2020-nile}, which complicates their integration with causal inference. \citet{zhang2023towards} propose two casual metrics to identify valid rationales when two or more snippets are highly inter-correlated. 

\subsection{Continuous Prompts and  Explanations}
Continuous prompts refer to continuous vectors incorporated into the input of a prompted language model. These prompts are typically tuned using gradient-based methods, which are directed by the labeled training examples of the tasks \cite{qin-eisner-2021-learning, suvcik2023prompterator, huang2023event,shah2023adept,liu2023gpt}. Although these prompts generally enhance model performance, their continuous nature poses challenges for human understanding and interpretation \cite{work0,work1,work2,work3}. \citet{work1} test whether few-shot prompt-based models depend on shallow cues, the results reveal that simple shortcuts still exist. \citet{work2} use adversarial prompts to inject knowledge into models, outperforming manually designed prompts. \citet{work3} propose to use discrete words to explain the continuous prompts in a bag-of-words manner. The interpretations, as shown in Figure \ref{fig:intro1}, made up of words  sorted by the learned weights, exhibit fragmented semantics, which impairs plausibility. 
% \section{Submodular Optimization}

\section{Start From Attention }
In this section, we give a brief formulation on how the continuous prompts can be decomposed into semantically related concepts.
% As \citet{ferrando-etal-2022-measuring} mentioned, the attention block  can be reformulated into a combination of the layer input representations. 
Given a prompt augmented sequence of token representations $\boldsymbol{T}=(\boldsymbol{t}_1,\cdots,\boldsymbol{t}_n)\in\boldsymbol{R}^{d\times n}$, and a model with $H$ heads. $d$ is the embedding dim and $d_h=d/H$. For simplicity, we suppose the continuous token is the $i$-th token.
Each Multi-Head Attention (MHA) computes $\boldsymbol{z}_i^h$ as follows:
\begin{equation}
    \boldsymbol{z}_i^h=\sum_j\boldsymbol{A}_{i,j}^h\boldsymbol{W}_{V}^h\boldsymbol{t}_j
\end{equation}
with $\boldsymbol{A}_{i,j}^h$ referring to the attention weight where token $i$ attends token $j$, and $\boldsymbol{W}_{V}^{h}\in\mathbb{R}^{d_h\times d}$ is the weight matrix of $h$-th head. 

% As the Layer Normalization (LN) can be reformulated into a linear transformation operation, we only need to consider the residual connection \cite{ferrando-etal-2022-measuring}.

Generally, the FFN block has two layers, and we use $\boldsymbol{r}_{m}$, $\boldsymbol{p}_{m}\in\mathbb{R}^{d}$ to denote the $m$-th column vectors of the inner weight matrix $\boldsymbol{W}_1\in\mathbb{R}^{d\times M}$ and  outer weight matrix $\boldsymbol{W}_2\in\mathbb{R}^{d\times M}$. Suppose the activation function is $\phi$. We can obtain the final output $\boldsymbol{y}_i$ as follows:
\begin{equation}
    \begin{split}
\boldsymbol{y}_i&=\sum_h\sum_j\sum_m\boldsymbol{A}_{i,j}^h\phi(\boldsymbol{r}_m^T\boldsymbol{W}_O^h\boldsymbol{W}_V^h\boldsymbol{t}_j)\boldsymbol{p}_{m}+\\
&\sum_h\sum_j\sum_m \phi(\boldsymbol{r}_m^T\boldsymbol{t}_j)\boldsymbol{p}_{m}
    \end{split}
\end{equation}
with $m$ referring the $m$-th dimension of inner hidden states. $\boldsymbol{W}_{O}^{h}\in\mathbb{R}^{d\times d_h}$ is the partitioned weight matrix. The second term is deduced by the residual connection trick. As the the Layer Normalization  can be reformulated into a linear transformation operation \cite{ferrando-etal-2022-measuring}, we only take the residual into consideration.

Notice that all the terms preceding \( \boldsymbol{p}_m \) are scalars, so the final output can be viewed as a linear combination of \( \boldsymbol{p}_m \). Based on findings of  \citet{transformers_key_value_memory}, each $\boldsymbol{p}_m$ corresponds to a shallow or semantic pattern, which can be described with concepts. Therefore, we decompose the continuous prompts into human-readable concepts.
\section{Problem Formulation}
Given a training set of text $\mathcal{X}=\{(x,y)_l\mid l\in|\mathcal{X}|\}$, where $x$ is the text and $y\in \mathcal{Y}$ represents a label from a set of $N$ classes. Normally we use $\mathcal{X}$ to tune the continuous prompts $\boldsymbol{P}\in\mathbb{R}^{d\times N_q}$ with $N_q$ being the number of continuous prompt tokens. Let $\mathcal{S}$ be the superset of candidate textual concepts generated from GPT-4o and $\mathcal{C}\subseteq \mathcal{S}$ be the desired set  with $N_c$ concepts. 

% we utilize a text encoder to extract text features from all texts in $\mathcal{X}$, denoted as $\mathcal{E}$.  We employ a submodular function $\mathcal{F}$ to select a concept set $\mathcal{C}$, where $\mathcal{C}\subseteq\mathcal{S}$ and consists of $N_c$ concepts.

We then initialize  concept embeddings $\boldsymbol{C}\in\mathbb{R}^{d\times N_c}$ and coefficient $\boldsymbol{Q}\in\mathbb{R}^{N_c\times N_q}$. Our goal is to decompose the continuous prompts into $\boldsymbol{CQ}$ and ensure the explanation fidelity \cite{NEURIPS2019_a7471fdc}.
% Given input matrix $\boldsymbol{X}\in\mathbb{R}^{d\times n}$ and continuous prompt matrix $\boldsymbol{P}\in\mathbb{R}^{d\times N_q}$, we first analyze whether there exists a concept matrix $\boldsymbol{C}\in \mathbb{R}^{d\times N_c}$ and coefficient matrix $\boldsymbol{Q}\in\mathbb{R}^{N_c\times N_q}$ ($N_c$ and $N_q$ denote the number of concepts and prompts respectively).

% enabling us to interpret $\boldsymbol{P}$ from both fidelity and sensitivity perspectives \cite{NEURIPS2019_a7471fdc}.
% First we give the definitions of fidelity and sensitivity of interpreting continuous prompts to define what is a good explanation.
% Given a black-box model $\theta$, explanation function $\Phi$, we define the explanation fidelity of $\Phi$ as:
% \begin{equation}
%     Fi(\Phi) = \mathbb{E}[d(P(y|\theta,\boldsymbol{X},\boldsymbol{P}),\Phi(\boldsymbol{X},\boldsymbol{CQ}))] 
% \end{equation}
% with $d$ represents distance between two outcome.
% Given binary random variable $\boldsymbol{I}\in \mathbb{R}^{N_q}$,
% we define the explanation sensitivity as:
% \begin{equation}
%     Sens(\Phi)=\mathbb{E}_{\boldsymbol{I}}[d(P(y|\boldsymbol{X},\boldsymbol{CQ}), \Phi(\boldsymbol{X},\boldsymbol{CQI}))]
% \end{equation}
Now we give the following  lemma to ensure the decomposition is feasible\footnote{The proof is available in Appendix \ref{sec:proof}.}.

\noindent\textbf{{Lemma}} For any $\boldsymbol{P}\in\mathbb{R}^{d\times N_q}$, $\epsilon>0$, there exist $\boldsymbol{C}\in\mathbb{R}^{
d\times N_c}$, $\boldsymbol{Q}\in\mathbb{R}^{N_c\times N_q}$ that satisfies $||\boldsymbol{CQ}-\boldsymbol{P}||_F^2\leq \epsilon$.

With this lemma, we can assure that for each continuous prompts, a corresponding concept can be identified.
\section{Methodology}
\begin{figure*}[!htb]
    
    % \subfigure[The framework.]{
        % \begin{minipage}{\textwidth}
            \centering
            \includegraphics[width=\textwidth]{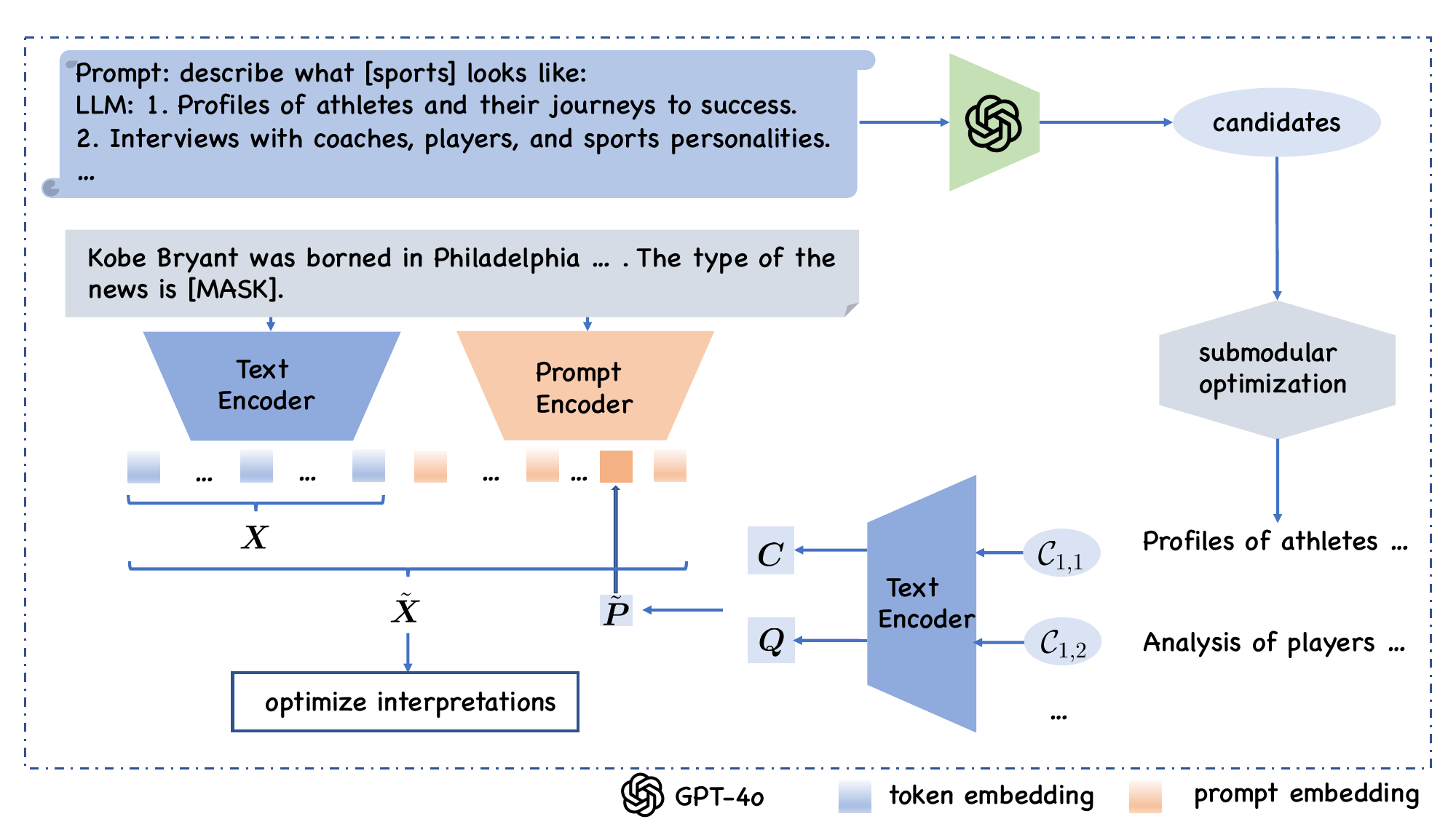} 
            \caption{Overview of our framework CD. Our framework prompts GPT-4o to generate candidate concepts for each class (Section \ref{step1}). We then use submodular optimization to select ones that maximize diversity  and coverage (Section \ref{step2}). Next, we initialize the concept embedding \(\boldsymbol{C}\) by feeding the concepts into a text encoder (BERT-Large or GPT2-Medium) and initialize the coefficient embedding \(\boldsymbol{Q}\). Finally, we tune the embeddings using stochastic gradient descent (Section \ref{step3}). $\tilde{\boldsymbol{X}}$ denotes the prompt-augmented input. $\tilde{\boldsymbol{P}}=\boldsymbol{CQ}$.           }
        %     \label{fig:framework1}
        % \end{minipage}
        % \enquote{filter} represents the filtering process in Section \ref{step3}.  
    % }
\label{fig:framework}

\end{figure*}
As illustrated in Figure \ref{fig:framework}, our framework prompts  GPT-4o  to generate a set of candidate concepts for each class (Section \ref{step1}). We employ submodular optimization to greedily select a subset of concepts for each class such that we maximize  diversity and coverage scores (Section \ref{step2}). We then initialize the concept embedding $\boldsymbol{C}$ by feeding the concepts into a text encoder  and the coefficient embedding $\boldsymbol{Q}$. Ultimately we design a loss to tune the two embeddings (Section \ref{step3}).

\subsection{Generate the Candidates}
\label{step1}
We first generate a candidate set of concepts by combining prompts from labeled text. To ensure text diversity, we generate 50 samples and design corresponding prompts for the experimental dataset, as shown in Table \ref{table1}. Universal prompt templates need to adapt to specific information in the dataset. For instance, in the IMDB dataset \cite{maas-etal-2011-learning}, template \ref{table1} will be modified to \enquote{describe what a positive review looks like}.
Additionally, due to the uncontrollability of LLMs \cite{bhargava2024whats}, we append  auxiliary instructions following prompts to avoid label leakage. However, even with these precautions, we still find some text instances containing labels. Therefore, we remove parts of the generated text through regular expression matching.

\begingroup
\setlength{\tabcolsep}{1pt} 
\begin{table}

\begin{tabular}{l}
\hline Universal Prompt Template \\
\hline 1. describe what a [CLASS NAME] looks like: \\
2. describe the aspects of a [CLASS NAME]: \\
3. describe [CLASS NAME] using  sentences:\\
4. describe the patterns of a [CLASS NAME]: \\
5. describe the concepts of a [CLASS NAME]: \\
\hline
\end{tabular}
\caption{The prompt templates used to generate raw sentences from GPT-4o.}
\label{table1}
\end{table}

\subsection{Submodular Optimization}
% We generate a superset of candidate concepts, denoted as $S$, from class-specific subsets. For each label $y\in Y$, we build $S_y$ by prompting a language model to generate textual knowledge about $y$. Rather than directly choosing $N_c$ concepts from $S$, we opt to select $k$ concepts for each class, ensuring that $N \times k = N_c$, thus guaranteeing an equal distribution of relevant concepts for each class in the bottleneck.

To ensure that each category contributes equally to each concept, we need to select $k$ relevant concepts for each category $y$. We   use the corresponding prompts from Table \ref{table1} to generate the overall candidate set $\mathcal{S}$, and we split $\mathcal{S}$ for each category as $\mathcal{S}_y$.

The next step is to select the desired set $\mathcal{C}_y$ from $\mathcal{S}_y$, which satisfies two properties: diversity and coverage. Diversity means that concepts are more informative, while the second requires good coverage of the candidate set. This problem is a typical combinatorial optimization problem. Fortunately, we can solve it in polynomial time complexity using submodular optimization \cite{submodularnips}. We can measure the two properties above with a function $\mathcal{F}:2^{|\mathcal{C}_y|}\rightarrow \mathbb{R}$. A submodular function  satisfies the diminishing property\footnote{A set function $\mathcal{F}$ satisfying diminishing property is that for every $\mathcal{A}\subseteq\mathcal{B}$ and $x\notin \mathcal{B}$, $\mathcal{F}(\mathcal{A}\cup\{x\})-\mathcal{F}(\mathcal{A})\geq \mathcal{F}(\mathcal{B}\cup\{x\})-\mathcal{F}(\mathcal{B}).$}. If the diminishing function is monotonic\footnote{A monotonic submodular function $\mathcal{F}$ is that for every $\mathcal{A}\subseteq\mathcal{B}, \mathcal{F}(\mathcal{A})\leq\mathcal{F}(\mathcal{B}).$}, then the problem can be solved using a greedy algorithm \cite{Nemhauser_Wolsey_Fisher_1978}. Inspired by \citet{submodular3}, we propose the following monotonic submodular function to select $\mathcal{C}_y$ from $\mathcal{S}_y$:
\begin{equation}
    \mathcal{F}(\mathcal{C}_y)=\lambda\mathcal{L}_d(\mathcal{C}_y)+\mathcal{L}_c(\mathcal{C}_y)
    \label{sub_func}
\end{equation}
with $\mathcal{L}_d(\mathcal{C}_y)$  evaluates diversity and $\mathcal{L}_c(\mathcal{C}_y)$ rewards coverage. The first term would give a higher score when containing  informative concepts, and the second term is higher when concepts are representative. The hyperparameter $\lambda$ controls the weights of two terms.

\noindent\textbf{Diversity Score}  We propose the following diversity score:
\begin{equation}
    \mathcal{L}_d(\mathcal{C}_y)=\sum_{c\in \mathcal{C}_y}\psi\left(\sum_{y\in \mathcal{Y}}Div(y,c)\right)
\end{equation}
with $\psi$ is the concave function and $Div(y,c)$ measures the differences between $\mathcal{C}_y$ and $\mathcal{Y}$. Here, we use a simple affine function (i.e. $\psi(x)=x$). And we calculate $Div(y,c)$ as follows:
\begin{equation}
    Div(y,c)=\sum_{y\in \mathcal{Y}}Sim(y|c)\cdot \text{log}\left(Sim(y|c)\right)
\end{equation}
\begin{equation}
    Sim(y,c)=\frac{1}{|\mathcal{X}_y|}\sum_{x\in \mathcal{X}_y}\mathcal{E}(x)\cdot \mathcal{E}(c)^T 
\end{equation}
with $\mathcal{E}(\cdot)$ represents the encoder.

\noindent\textbf{Coverage Score} We define the following coverage score:
\begin{equation}
    \mathcal{L}_c(\mathcal{C}_y)=\sum_{c_1\in \mathcal{S}_y}\sum_{c_2\in \mathcal{C}_y}\phi(c_1,c_2)
\end{equation}
where $\phi$ represents the similarity measure between $c_1$ and $c_2$. This score is a minimax facility location metric \cite{aus}, aiming to minimize the maximum distance between each element in the subset and the candidate set. The distance measure employed is the cosine similarity between the features of the two concepts  extracted by the text encoder. A high coverage score indicates a diverse concept encompassing a target class's potential aspects.
\label{step2}

\subsection{Optimize Interpretations}
\label{step3}

In this section, we introduce the method for solving $\boldsymbol{C}$ and $\boldsymbol{Q}$. A classic post-hoc explanation system derives the corresponding explanation based on the input and output \cite{post-hoc}. We require our explanations to stay consistent with original continuous prompts while maintaining the performance. Based on prior studies \cite{NEURIPS2019_a7471fdc, work3}, we construct two loss terms to meet these requirements.  For mitigating deviation, we introduce the following constraint:
\begin{equation}
    \mathcal{L}_{f}=\mathcal{D}_{\text{KL}}(\mathcal{P}(y|\boldsymbol{X},\boldsymbol{P})\parallel \mathcal{P}(y|\boldsymbol{X},\boldsymbol{CQ}))
\end{equation}
where $\mathcal{P}(\cdot)$ represents the output probability.
% This KL penalty requires that $f(\boldsymbol{X},\boldsymbol{P})$ does not deviate excessively from $f(\boldsymbol{X},\boldsymbol{CQ})$.
For performance retention, we introduce the following constraint:
\begin{equation}
    \mathcal{L}_{l}=-\text{log}\mathcal{P}(y|\boldsymbol{X},\boldsymbol{CQ}).
\end{equation}
% This constraint measures the difference between replacing $\boldsymbol{P}$ with $\boldsymbol{CQ}$ from the perspective of correlation. 
% A smaller difference indicates higher loyalty.
We combine these two terms in the following manner:
\begin{equation}
    \mathcal{L}=\mu\mathcal{L}_{f}+\mathcal{L}_{l}
    \label{loss}
\end{equation}
with $\mu\in[0,1]$. 
% Each training step we   
\subsection{Local Level Interpretations}
In this section, we deliberately introduce our strategy for generating explanations. Explaining a continuous prompt is  non-trivial. The difficulty lies in the fact that a global explanation may not be plausible enough \cite{work3}. We are more concerned with why a continuous prompt model makes a particular decision given the input and output (i.e. post-hoc manner). 

Therefore, we propose the following local level explanation strategy. For input $(x,y)_l$,  we optimize  $\mathcal{F}(\mathcal{C}_y)$ (\ref{sub_func}) to obtain candidates. Then we select top-k concepts as our explainable concepts sorted by keys derived from $\boldsymbol{Q}$. The sorting key of $c_{y,p}$ is calculated as:
\begin{equation}
    \text{key}(c_{y,p})=\sum_{q=0}^{N_q-1}\boldsymbol{Q}_{p,q}
\end{equation}
where $c_{y,p}$ denotes the $p$-th element of $\mathcal{C}_y$.

\section{Experiments}
 
\subsection{Experimental Setups}
\noindent\textbf{Metrics and Baselines}\quad We investigate the interpretability of CD by comparing it to other explanation methods using two metrics: accuracy and concept correlation. Following \citet{work3}, we use accuracy to assess how well the concept prompt aligns with the continuous prompts. We adopt Vocab-1500 explanation methods as our baseline, with 1500 being the maximum number of vocabularies according to \citet{work3}. Inspired by \citet{ig_v1}, we introduce a new metric called Concept Correlation, denoted as $\rho$, to evaluate concept-level explanations. Concretely, we compute the Pearson correlation between the coefficients learned by a bag-of-concepts model and the attribution scores provided by various explanation methods. For   baselines, we choose three  attribution methods to assign scores for each prompt. These scores are comparable to coefficients for explanation faithfulness. Specifically 
we choose TokenSHAP \cite{tokenshap}, IG \cite{sundararajan2017axiomatic}, and grad \cite{grad*shap}. 

\noindent\textbf{Datasets and Backbones}\quad Throughout our experiments, we primarily study BERT-Large (335M) and GPT-2-Medium (345M) on text classification tasks. In Section \ref{general_results}, we utilize the SST-2 \cite{socher-etal-2013-recursive}, IMDB \cite{maas-etal-2011-learning}, and AGNews \cite{NIPS2015_250cf8b5} datasets.  

\noindent\textbf{Implementation Details}\quad We implement both full and few-shot learning settings  (i.e. 4-shot, 8-shot, 16-shot, 32-shot, and full-shot). We design two different types of prompts for each dataset and customize them to suit diverse datasets\footnote{Appendix \ref{more_results} provides results with more types of prompts.}. Throughout our experiments, we implement four different prompt length settings, including 1, 2, 5, 10. Following \citet{liu2023gpt}, we train continuous prompts and concept embeddings using the AdamW optimizer with a learning rate of $10^{-4}$. We employ early stopping with 100 iteration steps. Additional hyperparameters are provided in the Appendix. Results are reported using five random seeds. 
 
\begingroup
\setlength{\tabcolsep}{0.01pt} % Default value: 6pt
\renewcommand{\arraystretch}{1.1}
\begin{table*}[!htbp]
\centering
\begin{tabular}{llcccccccccc}
\toprule[1.2pt]
\multicolumn{2}{c}{\multirow{3}{*}{}} & \multicolumn{6}{c}{BERT-Large}                      & \multicolumn{4}{c}{GPT-2-Medium}                     \\ 
% \cline{3-12} 
\multicolumn{2}{c}{}                  & \multicolumn{3}{c}{$N_q=1$}  & \multicolumn{3}{c}{$N_q=2$}  & \multicolumn{2}{c}{$N_q=1$}  & \multicolumn{2}{c}{$N_q=2$}  \\ 
\cmidrule(lr){3-5}\cmidrule(lr){6-8}\cmidrule(lr){9-10}\cmidrule(lr){11-12}
\multicolumn{2}{l}{Dataset \, shot}                  & P-tuning & CD & Vocab & P-tuning & CD & Vocab & P-tuning & CD  & P-tuning & CD  \\ \Xhline{1.2pt}
\multirow{6}{*}[-1.5ex]{ SST-2}    & 4    &  0.581        &  0.592  &    0.556      &   0.625       & 0.606  &  0.620        &   {0.560}       & 0.557   & 0.533         & {0.545}                  \\ \cline{2-12} 
                          & 8    &  {0.691}        &  0.615  & 0.618         &  0.579        & {0.674}   &  0.642        &   {0.636}       & 0.596   &  {0.605}        &  0.603                  \\ \cline{2-12} 
                          & 16   &   {0.840}       & 0.797   & 0.818         &  0.744        & 0.744   &   {0.752}       &  0.831        & {0.848}   &   {0.821}       & 0.809                   \\ \cline{2-12} 
                          & 32   &  0.847        & 0.850   &{0.859}          & {0.831}         & 0.812   &   0.782       &   0.839       & {0.852}   &  0.832        &  {0.856}                  \\ \cline{2-12} 
                          & Full      &   {0.910}       & 0.907   & 0.907         &  0.896        & 0.892   &  {0.904}        &  0.905        & {0.917}   & {0.906}         &  0.840                  \\ \cline{2-12} 
                          & Avg & \textbf{0.773} & 0.752& 0.751& 0.735& \textbf{0.745}& 0.740&0.754&0.754&\textbf{0.739}&0.730\\ \cline{2-12}
                          & ${\sigma^2}$       &  ${\pm0.028}$        & ${\pm0.005}$   &  ${\pm0.003}$        &  $\pm0.002$        &  $\pm0.003$  & $\pm0.004$         &   $\pm0.0004$       &  $\pm0.0005$  &    $\pm0.0008$      &     $\pm0.003$               \\ \Xhline{1.2pt}
\multirow{6}{*}[-1.5ex]{ IMDB}     & 4    &  0.542        &  {0.577}  &    0.556      &   0.563       & 0.538   & {0.569}         &  0.547        &  {0.556}  &  0.529        &  0.529               \\ \cline{2-12} 
                          & 8    & 0.750         & {0.789}   &   0.688       &  {0.605}        & 0.549   &    0.536      & {0.509}         & 0.507   &  {0.518}       & 0.501                 \\ \cline{2-12} 
                          & 16   & 0.834         & {0.876}   &    0.852      & {0.848}         & 0.772   &   0.759       &  0.506        & {0.507}   & 0.504         & {0.527}                   \\ \cline{2-12} 
                          & 32   & 0.839         & 0.848   &     {0.867}     &  {0.804 }       & 0.800   &  0.800        &  {0.790}        &0.714    &  0.763        &  {0.864}                  \\ \cline{2-12} 
                          & Full      & 0.918         & 0.917   &   {0.919}       & {0.919}         & 0.916   &   0.917       & 0.924         &{0.926}    &  0.909        &  {0.919}                  \\ \cline{2-12} 
                          & Avg & 0.776 & \textbf{0.801}& 0.776&\textbf{0.747}&0.715&0.716&\textbf{0.655}&0.642&0.644&\textbf{0.668}\\ \cline{2-12}
                          & $\sigma^2$       &  $\pm0.004$        & $\pm0.0009$   &   $\pm0.002$       & $\pm0.001$         & $\pm0.003$   &    $\pm0.004$      &  $\pm0.0004$        &  $\pm0.001$  &  $\pm0.001$        &  $\pm0.0005$                \\ \Xhline{1.2pt}
\multirow{6}{*}[-1.5ex]{ AGNews}   & 4    &  0.687        &0.704    & {0.712}         &  0.683        & {0.698}   & 0.663         &  {0.762}        &0.755    &  0.759        &  0.762                  \\ \cline{2-12} 
                          & 8   &  {0.742}        &0.692    &  0.689        & 0.712         &  0.711  & {0.741}         & 0.790         & 0.790   &  {0.797}        &  0.796                  \\ \cline{2-12} 
                          & 16   &  {0.767}        &  0.766  &  0.758        & {0.759}         & 0.755   & 0.747         & 0.784         &{0.793}    &  0.781        & {0.804}                \\ \cline{2-12} 
                          & 32   & 0.823         &  0.814  &  {0.828}        &  0.828        & {0.837}   &  0.805        & 0.836         & {0.837}   & 0.831         & 0.831                 \\ \cline{2-12} 
                          & Full      &  0.882        &0.880    &  {0.884 }       & {0.888}         & 0.887   &  0.885        & 0.865         & {0.866}   &0.854          & {0.858}                  \\ \cline{2-12} 
                          & Avg & \textbf{0.780}& 0.771&0.774&0.774&\textbf{0.777}&0.768&0.807&\textbf{0.808}&0.804&\textbf{0.810}\\ \cline{2-12}
                          & $\sigma^2$       &  $\pm0.0009$        & $\pm0.0006$   &  $\pm0.001$        &  $\pm0.0006$        & $\pm0.0007$   &  $\pm0.001$         &  $\pm0.00003$        & $\pm0.00005$   &  $\pm0.0001$        &     $\pm0.0001$              \\ \Xhline{1.2pt}
\end{tabular}
\caption{Accuracy results of CD  with baselines on the SST-2, IMDB and AGNews dataset. $N_q$ denotes the number of continuous prompt tokens. Vocab represents Vocab-1500. $\sigma^2$ denotes the average variance across five shot settings.}
% \footnote{}
\label{acc_table}
\end{table*}

\begingroup
\setlength{\tabcolsep}{3.5pt}
\renewcommand{\arraystretch}{1.0}
\begin{table*}[!htbp]
\centering
\begin{tabular}{llcccccccc}
\Xhline{1.2pt}
\multicolumn{2}{c}{\multirow{2}{*}{}} & \multicolumn{4}{c}{BERT-Large} & \multicolumn{4}{c}{GPT-2-Medium} \\
\multicolumn{2}{c}{Dataset\,top-k}           &     CD &  TokenSHAP   & IG  & Grad  & CD   & TokenSHAP   & IG   & Grad   \\ \Xhline{1.2pt}
\multirow{4}{*}[-1.5ex]{ SST-2}     & 3    & {0.755}     &   0.646        & 0.525    & 0.481      &  {0.824}    &  0.613         &0.535     &   0.342     \\ \cline{2-10} 
                           & 5    & {0.561}     &      0.427     & 0.221    & 0.264      & {0.610}     & 0.497          &  0.353    &0.289         \\ \cline{2-10} 
                           & 10   &  {0.484}    &   0.398        & 0.141    & 0.173      &{0.445}      &0.310           & 0.252     & 0.200        \\ \cline{2-10} 
                           & Avg & \textbf{0.600}& 0.490&0.295&0.306&\textbf{0.626}&0.473&0.380&0.277\\ \cline{2-10}
                           & $\sigma^2$      &$\pm0.046$      &   $\pm0.034$        & $\pm0.062$    &  $\pm0.038$     & $\pm0.062$     &  $\pm0.041$         & $\pm0.017$     &  $\pm0.029$       \\ \Xhline{1.2pt}
\multirow{4}{*}[-1.5ex]{IMDB}      & 3    &     0.553 &   0.528        &  {0.669}   & 0.443      &  {0.824}    & 0.557          &0.466      &  0.382      \\ \cline{2-10} 
                           & 5    & {0.654}     &0.535           & 0.429    &0.294       & {0.610}     &0.476          & 0.302     &0.235         \\ \cline{2-10} 
                           & 10   &{0.378}     & 0.203          & 0.184    &0.217       & {0.445}     &  0.310         &0.220      & 0.187      \\ \cline{2-10} 
                           & Avg & \textbf{0.528}& 0.422&0.427&0.318&\textbf{0.626}&0.447&0.329&0.268\\ \cline{2-10}
                           & $\sigma^2$      &$\pm0.039$      &  $\pm0.025$         &    $\pm0.038$ & $\pm0.028$      & $\pm0.062$     & $\pm0.026$          &$\pm0.024$      &$\pm0.033$        \\ \Xhline{1.2pt}
\multirow{4}{*}[-1.5ex ]{AGNews}    & 3    &    0.553  & 0.505          &{0.629}     &0.557       & {0.824}     &0.513           &0.299      &     0.432   \\ \cline{2-10} 
                           & 5    &{0.654}      &0.454          &0.424     &   0.333    & {0.610}     &0.496           & 0.320     & 0.297       \\ \cline{2-10} 
                           & 10   & {0.378}     &   0.219        &  0.203   & 0.155      &  {0.445}    &   0.217        &0.156      &0.309        \\ \cline{2-10} 
                           &Avg&\textbf{0.528}&0.392&0.418&0.348&\textbf{0.626}&0.408&0.258&0.346\\ \cline{2-10}
                           & $\sigma^2$      &$\pm0.059$       &$\pm0.034$          &$\pm0.047$     & $\pm0.029$      &  $\pm0.059$    &$\pm0.077$          & $\pm0.161$     &  $\pm0.027$     \\ \Xhline{1.2pt}
\end{tabular}
\caption{Concept correlation results of CD with baselines.  $\sigma^2$ denotes the average variance across three top-k settings.}
% \footnote{}
\label{corr_table}
\end{table*}

\begingroup
\setlength{\tabcolsep}{1pt} % Default value: 6pt
\begin{table*}[]
\centering
\begin{tabular}{lll}
\Xhline{1.2pt}
Class  & Input & Top-3 Concepts \\ \hline

 world      & \emph{\makecell[l]{euters china has said \\no date has been set \\for working level talks\\ on the north korean \\nuclear crisis ...}}      &  \makecell[l]{1.\emph{Breaking news updates on significant global events.}\\2.\emph{In-depth analysis of international conflicts and resolutions.}\\3.\emph{Coverage of diplomatic relations between countries.}}              \\ \cline{1-3} 
                         sports     & \emph{\makecell[l]{michael phelps won \\the gold medal in \\the 400 individual \\medley ...}}      &  \makecell[l]{1.\emph{Live coverage and updates of ongoing sporting events.}\\2.\emph{In-depth analysis of match statistics and player performances.}\\3.\emph{Profiles of athletes and their journeys to success.}}              \\ \cline{1-3} 
                         business   & \emph{\makecell[l]{new york reuters u s \\treasury debt prices \\slipped on monday ...}} & \makecell[l]{1.\emph{In-depth analysis of stock market trends and movements.}\\2.\emph{Coverage of corporate earnings reports and financial performance.}\\3.\emph{Updates on global economic indicators such as GDP and inflation.}}
                        \\ \cline{1-3} 
                         sci/tech   & \emph{\makecell[l]{a company ... won a \\grant to develop \\a method of producing \\better peptides ...}} 
      &  \makecell[l]{1.\emph{Breaking news on scientific discoveries and breakthroughs.}\\2.\emph{In-depth analysis of emerging technologies and their implications.}\\3.\emph{Profiles of leading scientists, engineers, and innovators.}}          \\ \Xhline{1.2pt}
\end{tabular}
\caption{Explanation examples on  AGNews dataset with BERT-Large. Top-3 concepts are ranked based on the coefficients in $\boldsymbol{Q}$.}
\label{result_sst2_AGNews_bert}
\end{table*}
\endgroup

% \begingroup
% \setlength{\tabcolsep}{1.9pt} % Default value: 6pt
% % \renewcommand{\arraystretch}{1}
% \begin{table*}[]
% \centering
% \begin{tabular}{lll}
% \Xhline{1.2pt}
% Class  & Input & Top-3 Concepts \\ \hline
%  positive   & \emph{I loved it!}      &  \makecell[l]{1.\emph{Encouraging and Supportive.}\\2.\emph{Happy and Optimistic.}\\3.\emph{Constructive and Helpful.}}               \\ \cline{1-3} 
%                          negative   &  \makecell[l]{\emph{extremely}\\ \emph{confusing}}     &    \makecell[l]{1.\emph{Opposing or contradictory to desired outcomes.}\\2.\emph{Pessimistic outlook or attitude.}\\3.\emph{Discouraging or disheartening experiences.}}                 \\ \Xhline{1.2pt}

% \end{tabular}
% \caption{Explanation examples on SST-2 and AGNews dataset with BERT-Large.}
% \label{result_sst2_AGNews_bert}
% \end{table*}
% \endgroup
\subsection{General Results}
\label{general_results}
We first evaluate CD using accuracy metric across three datasets, as shown in Table \ref{acc_table}. In summary, we can draw following two conclusions: i) CD could reach a similar level to P-tuning (e.g. BERT-Large/$N_q=1$, P-tuning: 
 0.773, CD: 0.752), which demonstrates our effectiveness in fitting continuous prompts with two typical  language model architectures, further reflecting fidelity. The subtle differences demonstrate the feasibility of our framework. 
ii) CD, with  concepts whose average lengths reach up to 300, can perform similarly to the discrete explanation method with 1500 vocabulary capacity (e.g. BERT-Large/$N_q=2$, CD:0.715, Discrete:0.716). Moreover, in some settings outperforms Discrete (e.g. BERT-Large/IMDB/$N_q=1$, CD: 0.801, Discrete: 0.776).  

We then conduct correlation experiments on these three datasets with the full shot setting, and the results are shown in Table \ref{corr_table}. We find that: i) In 7 out of 9 experimental settings, we achieved the best results, indicating that our method is faithful to continuous prompts. ii) As the number of concepts increases, the correlation decreases rapidly, which suggests that our explanations still suffer from noise issues. iii) We test different continuous prompt tokens and find that no significant difference, which demonstrates the generalizability of out method. We conduct more depth analysis about trade offs between interpretability and performance of CD in Section \ref{trade_appendix}.

\begin{figure}[tbp]
    \centering
    \includegraphics[width=0.45\textwidth]{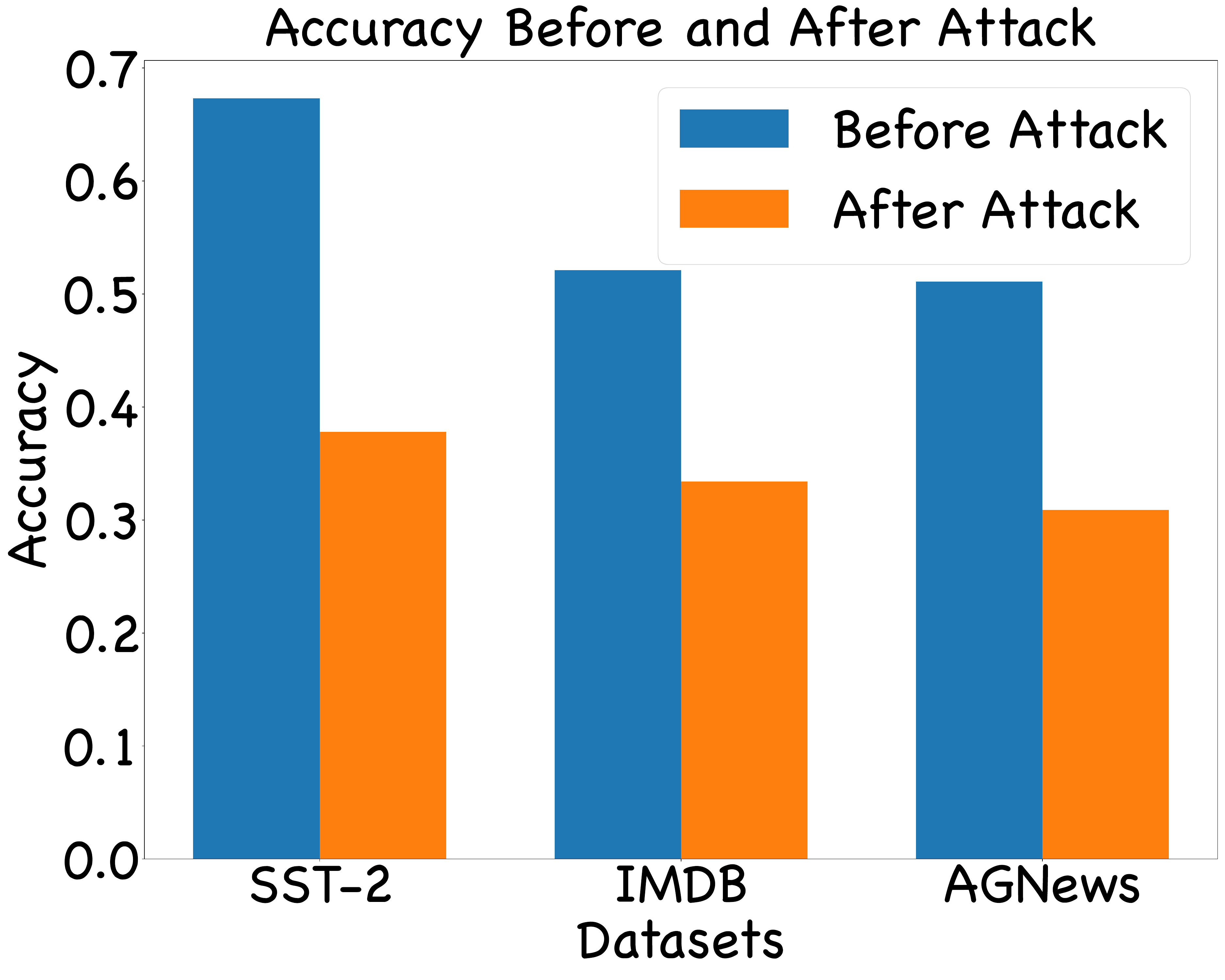}
    \caption{Comparison results before and after attack on SST-2, IMDB and AGNews datasets on BERT-Large.}
    \label{fig:attack}
\end{figure}

\begingroup
\setlength{\tabcolsep}{1pt} % Default value: 6pt
\renewcommand{\arraystretch}{1.1}
\begin{table*}[tbp]
\centering
\begin{tabular}{lcccccccccc}
\toprule[1.2pt]
\multicolumn{1}{c}{\multirow{3}{*}{}} & \multicolumn{6}{c}{BERT-Large}                      & \multicolumn{4}{c}{GPT-2-Medium}                     \\ 
% \cline{3-12} 
\multicolumn{1}{c}{}                  & \multicolumn{3}{c}{$N_q=5$}  & \multicolumn{3}{c}{$N_q=10$}  & \multicolumn{2}{c}{$N_q=5$}  & \multicolumn{2}{c}{$N_q=10$}  \\ 
\cmidrule(lr){2-4}\cmidrule(lr){5-7}\cmidrule(lr){8-9}\cmidrule(lr){10-11}
\multicolumn{1}{c}{}                  & P-tuning & CD & Discrete & P-tuning & CD & Discrete & P-tuning & CD  & P-tuning & CD  \\ \Xhline{1.2pt}
 4-shot    &  0.519        &  0.588  &    0.533      &   0.543       & 0.514  &  0.592        &   {0.551}       & 0.542   & 0.542         & {0.553}                  \\ \cline{1-11} 
                           8-shot    &  {0.514}        &  0.525  & 0.563         &  0.583        & {0.615}   &  0.524        &   {0.620}       & 0.597   &  {0.628}        &  0.583                  \\ \cline{1-11} 
                           16-shot   &   {0.646}       & 0.773   & 0.682         &  0.746        & 0.797   &   {0.790}       &  0.792        & {0.804}   &   {0.769}       & 0.772                   \\ \cline{1-11} 
                           32-shot   &  0.776        & 0.833   &{0.795}          & {0.793}         & 0.850   &   0.804       &   0.822       & {0.823}   &  0.814        &  {0.841}                  \\ \cline{1-11} 
                           Full      &   {0.868}       & 0.897   & 0.874         &  0.909        & 0.883   &  {0.869}        &  0.894        & {0.906}   & {0.894}         &  0.908                  \\ \cline{1-11} 
                           Avg & {0.664} & \textbf{0.723}& 0.689& 0.714& \textbf{0.731}& 0.715&0.735&0.734&{0.729}&0.731\\ \cline{1-11}
                           ${\sigma^2}$       &  ${\pm0.017}$        & ${\pm0.009}$   &  ${\pm0.005}$        &  $\pm0.001$        &  $\pm0.007$  & $\pm0.008$         &   $\pm0.0009$       &  $\pm0.0007$  &    $\pm0.0006$      &     $\pm0.004$               \\ \Xhline{1.2pt}

\end{tabular}
\caption{Accuracy results of CD  with baselines on the SST-2 dataset. $\sigma^2$ denotes the average variance across five shot settings.}
% \footnote{}
\label{acc_table_long}
\end{table*}
\subsection{Interpretation Visualization}
\begingroup
\setlength{\tabcolsep}{1pt} % Default value: 6pt
\begin{table}[]
\centering
\small
\begin{tabular}{l}
\Xhline{1.2pt}
  Top-3 Concepts \\ \hline
       \makecell[l]{1.\emph{Breaking news on mergers and acquisitions.}\\2.\emph{In-depth analysis of stock market trends and movements.}\\3.\emph{Profiles of successful entrepreneurs and  leaders.}}  \\ 
 \Xhline{1.2pt}
\end{tabular}
\caption{One business news from AGNews dataset. The input is \emph{london reuters oil prices surged to a new high of 47 a barrel on wednesday after a new threat by rebel militia against iraqi oil facilities and as the united states said inflation had stayed in check despite rising energy costs}.  }
\label{bad_case_agnews}
\end{table}
\endgroup

According to results from Table \ref{corr_table}, we choose top-3 concepts as sample explanations because top-5 or top-10 concepts contain noises which decreases the correlations. The  examples are provided in the Table \ref{result_sst2_AGNews_bert}. As seen, our generated explanations align well with the input, resulting in credible explanations. 
% For the SST-2 dataset, we have the LLM generate descriptions of both good (positive) and bad (negative) movies, providing  reasons for each. For  AGNews, the contexts get longer and the  categories expand to 4, which means tougher for interpretation. 
% We prompt GPT-4o to describe diverse kinds of news and encourage it to give aspects as many as possible. The results show the operability. 

Nonetheless, there still exists some confusing explanations which the correspondence is not obvious. Take Table \ref{bad_case_agnews} as an example, the input is about oil price fluctuating while the first concept focuses on company merge.  One possible reason may be   that continuous prompts rely on  short cut links \cite{work1,work3}. We conjecture that one reason of that maybe short cut links, and we can test in this way: we choose bad cases predicted by CD. For given label $y$, we select label $y^{\prime}\neq y$, and we choose the top-$3$ concepts in $\mathcal{C}_{y^{\prime}}$ as the candidate concepts to generate continuous prompts instead of submodular optimization. We call one change of concepts as \enquote{causal-attack}. And we calculate the average cosine similarities between input and generated explanation. The results are showed in Figure \ref{fig:attack}. We can infer that one possible reason of "noise issue" is the short cut link.

\subsection{Longer $N_q$ Scenario}
\label{longer_scenario}
To test the performance when $N_q$ increases, we conduct experiments with $N_q=5$ and $N_q=10$ on SST-2 dataset. The results are reported in Table \ref{acc_table_long}. We can draw the conclusion that CD could apply to scenarios which $N_q$ are larger than $1$ and $2$.

\section{Conclusion}
In this paper, we introduce CD, a framework to interpret continuous prompts employing generated optimized concepts to decompose the prompt embeddings. Specifically, we prove that continuous prompts could be decomposed into concepts. Then we prompt GPT-4o to generate candidates and use a submodular optimization algorithm to select the concepts. Finally we design a specific loss tune the concept embeddings and propose a local-level strategy to generate explanations. We conduct extensive experiments on three datasets across two typical language model architectures with various prompt types.  Results demonstrate that the effectiveness of our framework. 
% \section*{Acknowledgements}
\section*{Limitations}
 Based on the correlation experiment results, the selected concepts still suffer from noise issue, in this study we only propose a simple heuristic strategy to filter irrelevant samples. How to efficiently choose vital concepts needs further exploration.

\bibliography{custom}

\appendix

\clearpage
\section{Proof}
\label{sec:proof}
\noindent\textbf{{Lemma}} For any $\boldsymbol{P}\in\mathbb{R}^{d\times N_q}$, $\epsilon>0$, there exist $\boldsymbol{C}\in\mathbb{R}^{
d\times N_c}$, $\boldsymbol{Q}\in\mathbb{R}^{N_c\times N_q}$ that satisfies $||\boldsymbol{CQ}-\boldsymbol{P}||_F^2\leq \epsilon$.
\begin{proof}\let\qed\relax
First, it is straightforward that if $h(\C)=\left\|\C\Q-\Po \right\|_F^2$ is convex, then we can obtain a $\Q^{*}$ such that $ \left\|\C\Q-\Po \right\|_F^2\leq \epsilon$ \cite{zinkevich2003online}. Therefore we only need to prove that $h(\C)$ is convex. For any $\C$,
\begin{equation}
    h(\C)=\left\|\C\Q-\Po \right\|_F^2=tr((\C\Q-\Po)^T(\C\Q-\Po)).
\end{equation}
To prove that the function $h(\C)$ is convex, we need to show that its Hessian matrix is positive semi-definite for all $\C$. Let's denote the $ij$-th element of the Hessian matrix as $\Ho_{ij}$. The Hessian matrix of $h(\C)$ is given by:
\begin{equation}
    \Ho_{ij}=\frac{\partial ^2 h}{\partial \C_{ij}\partial \C_{kl}}.
\end{equation}
To prove convexity, we need to show that for any vector $\boldsymbol{v}$, $\boldsymbol{v}^T\boldsymbol{Hv}\geq 0$.
\begin{equation}
    \Ho_{ij}=\frac{\partial^2 }{\partial \C_{ij}\partial \C_{kl}}(\left\|\C\Q-\Po\right\|_F^2).
\end{equation}
Using the properties of the Frobenius norm, we can rewrite 
\begin{equation}
    \begin{split}
        \Ho_{ij}&=\frac{\partial}{\partial \C_{ij}}(\frac{\partial}{\partial \C_{ij}}tr((\C\Q-\Po)^T(\C\Q-\Po)))\\
    &=\frac{\partial}{\partial \C_{ij}}(2(\Q^T(\C\Q-\Po))_{kl})\\
    &=2\frac{\partial}{\partial \C_{ij}}((\Q^T(\C\Q-\Po))_{kl})\\
    &=2((\Q^T\Q)_{ik}\delta_{jl}),
    \end{split}
\end{equation}
$\delta_{jl}$ is the Kronecker delta. Note that $\Q^T\Q$ is positive semi-definite, and $\delta_{jl} \geq 0$, so $\Ho_{ij}$ is positive semi-definite.

% We can also proof this lemma from the matrix view. Let's expand $h(\boldsymbol{C})$ as follows:
% \begin{equation}
%     \begin{split}
%         h(\boldsymbol{C})=tr(&\boldsymbol{Q}^T\boldsymbol{C}^T\boldsymbol{CQ}-\boldsymbol{Q}^T\boldsymbol{C}^T\boldsymbol{P}-\boldsymbol{P}^T\boldsymbol{CQ}+\\&\boldsymbol{P}^T\boldsymbol{P}).
%     \end{split}
% \end{equation}
% Note that $tr(\boldsymbol{A})=tr(\boldsymbol{A}^T)$ and $tr(\boldsymbol{AB})=tr(\boldsymbol{BA})$, we can get:
% \begin{equation}
%     h(\boldsymbol{C})=tr(\boldsymbol{Q}^T\boldsymbol{C}^T\boldsymbol{CQ})-2tr(\boldsymbol{Q}^T\boldsymbol{C}^T\boldsymbol{P}) + tr(\boldsymbol{P}^T\boldsymbol{P}).
% \end{equation}
% To find the Hessian, we differentiate $h(\boldsymbol{C})$ with respect to $\boldsymbol{C}$:
% \begin{equation}
%     \nabla_{\boldsymbol{C}} h(\boldsymbol{C})=2(\boldsymbol{CQ}-\boldsymbol{P})\boldsymbol{Q}^T.
% \end{equation}
% We only need to consider the term related to $\boldsymbol{C}$, first the quadratic term is:
% \begin{equation}
%     tr(\boldsymbol{Q}^T\boldsymbol{C}^T\boldsymbol{CQ})=tr(\boldsymbol{C}^T\boldsymbol{C}\boldsymbol{QQ}^T).
% \end{equation}
% Since $\boldsymbol{QQ}^T$ is semi-definite, the Hessian of this term is:
% \begin{equation}
%     \nabla_{\boldsymbol{C}}^2 tr(\boldsymbol{C}^T\boldsymbol{C}\boldsymbol{QQ}^T)=2\boldsymbol{QQ}^T.
% \end{equation}
% Then the Hessian $2\boldsymbol{QQ}^T$ is semi-definite. 
\end{proof}
% \section{Datasets}

\section{Implementation Details}
\label{sec::implementation_details}
In this work, all language models are implemented using Transformers. All our experiments are conducted on a single A800 GPU, and the results are reported using 5 random seeds (i.e. 1, 42, 100, 999, 1756).

For BERT-Large and GPT2-Medium checkpoints (including MedBERT, LEGAL-BERT and FinBERT), we download model weights from Huggingface Model Hub.

For Discrete-1500, due to the lack of resource code, we implement this method ourselves. Specifically, we compare the vocab embeddings with label embeddings (feeding into BERT-Large/GPT-2-Medium) and we select top-1500 tokens based on cosine similarities as our explanation vocabulary.

\section{Trade-off between Interpretability and Performance}
\label{trade_appendix}
We conduct an analysis on with how the number of concepts per class (i.e. $|\mathcal{C}_y|$) affect the accuracy performance. The results in BERT-Large ($|N_q|=2$) are stated in Table \ref{trade_bert}.
\begingroup
\setlength{\tabcolsep}{1pt} % Default value: 6pt
\begin{table}[tbp]
\centering
\small
\begin{tabular}{llllll}
\Xhline{1.2pt}
  Method & 1 & 5 & 10 & 15 & 20 \\ \hline
  SST-2 & 0.691 &0.653&\textbf{0.745}&0.637&0.539\\ \hline
  IMDB&0.688&0.633&\textbf{0.715}&0.614&0.556\\ \hline
  AGNews&0.658&0.629&\textbf{0.777}&0.601&0.594 \\ \hline
\end{tabular}
\caption{Accuracy results with variant $|\mathcal{C}_y|$ ($N_q=2$) on BERT-Large.}
\label{trade_bert}
\end{table}
\endgroup
The results in GPT-2-Medium ($N_q=2$) are given in Table \ref{trade_gpt}.
\begingroup
\setlength{\tabcolsep}{1pt} % Default value: 6pt
\begin{table}[tbp]
\centering
\small
\begin{tabular}{llllll}
\Xhline{1.2pt}
  Method & 1 & 5 & 10 & 15 & 20 \\ \hline
  SST-2 & 0.652 &0.667&\textbf{0.730}&0.619&0.547\\ \hline
  IMDB  &0.649&0.619&\textbf{0.715}&0.621&0.539\\ \hline
  AGNews&0.618&0.605&\textbf{0.777}&0.611&0.581 \\ \hline
\end{tabular}
\caption{Accuracy results with variant $|\mathcal{C}_y|$ ($N_q=2$) on GPT2-Medium.}
\label{trade_gpt}
\end{table}
\endgroup
% The warm up scheduler every 10 steps increases the learning rate. 
\section{Dataset Statistics}

The statistical information for mentioned six datasets  is shown in Table \ref{datasettable}.
\begingroup
\setlength{\tabcolsep}{6pt} 
\renewcommand{\arraystretch}{1.0}
\begin{table}[h!]
\begin{tabular}{lllll}
\Xhline{1.2pt}
Dataset         & Train/Dev/Test & C     & L   \\
\hline
SST-2            & 6920/872/1821  & 2 & 50  \\
\hline
IMDB            & 20K/5K/25K     & 2  & 500 \\
\hline
AGNews & 25K/2K/7K               & 4  & 500 \\
\hline
Medical Abstracts & 10K/1K/2.8K   & 5 & 500 \\ 
\hline
JUSTICE &2.5K/0.1K/1.8K &8  & 500 \\
\hline
Finance Sentiment & 3.8K/0.1K/0.9K & 3 & 500 \\
\Xhline{1.2pt}
% SNLI &          10K/2K/3K & 3 & 16502 & 50\\
% \hline
\end{tabular}
\caption{Statistics of three datasets. C: number of classes, L: average text length}
\label{datasettable}
\end{table}
\section{Prompt List}
\begingroup
\setlength{\tabcolsep}{6pt} 
\renewcommand{\arraystretch}{1.0}
\setlength{\arrayrulewidth}{0.8pt} % Set the default line width
\begin{table}[!htbp]
\begin{tabular}{l}
\Xhline{1.2pt}
 Prompts \\ 
\hline
 \makecell[l]{1.describe what a \emph{positive  movie review}  looks like\\2.describe the aspects of a \emph{positive  movie review}\\3.describe a \emph{positive movie review} using sentences\\4.describe the patterns of a \emph{positive  movie review}\\5.describe the concepts of a \emph{positive  movie review}\\6.describe what a \emph{negative  movie review}  looks like\\7.describe the aspects of a \emph{negative movie review}\\8.describe a \emph{negative  movie review} using sentences\\9.describe the patterns of a \emph{negative movie review}\\10.describe the concepts of a \emph{negative  movie review}}       \\ \Xhline{1.2pt}

\end{tabular}
\caption{Prompts to generate candidate concepts on SST-2 dataset.}
\label{candidate_sst2_table}
\end{table}
\begingroup
\setlength{\tabcolsep}{6pt} 
\renewcommand{\arraystretch}{1.0}
\begin{table}[!htbp]
\begin{tabular}{l}
\Xhline{1.2pt}
 Prompts \\ 
\hline
 \makecell[l]{1.describe what a \emph{positive  review}  looks like\\2.describe the aspects of a \emph{positive  review}\\3.describe a \emph{positive  review} using sentences\\4.describe the patterns of a \emph{positive  review}\\5.describe the concepts of a \emph{positive  review}\\6.describe what a \emph{negative  review } looks like\\7.describe the aspects of a \emph{negative review}\\8.describe a \emph{negative  review} using sentences\\9.describe the patterns of a \emph{negative  review}\\10.describe the concepts of a \emph{negative  review}}       \\ 
 \Xhline{1.2pt}
\end{tabular}
\caption{Prompts to generate candidate concepts on IMDB dataset.}
\label{candidate_imdb_table}
\end{table}
\begingroup
\setlength{\tabcolsep}{6pt} 
\renewcommand{\arraystretch}{1.0}
\begin{table}[!htbp]
\begin{tabular}{l}
\Xhline{1.2pt}
 Prompts \\ 
\hline
 \makecell[l]{1.describe what a \emph{world news} looks like\\2.describe the aspects of a \emph{world news}\\ 3.describe what a \emph{sports news} looks like\\4.describe the aspects of a \emph{sports news}\\
 5.describe what a \emph{business news} looks like
\\6.describe the aspects of a \emph{business news}\\
7.describe what a \emph{sci/tech news} looks like\\
8.describe the aspects of a \emph{sci/tech news}}       \\ 
 \Xhline{1.2pt}
\end{tabular}
\caption{Prompts to generate candidate concepts on AGNews dataset.}
\label{candidate_agnews_table}
\end{table}

\begingroup
\setlength{\tabcolsep}{6pt} 
\setlength{\arrayrulewidth}{1.0pt} % Set the default line width
\renewcommand{\arraystretch}{1.0}
\begin{table}[!htbp]
\begin{tabular}{l}
\Xhline{1.2pt}
 Prompts \\ 
\hline
 \makecell[l]{1.use some phrases to describe \emph{neoplasms}
 \\2.use some sentences  to describe \emph{neoplasms}
 \\ 3.use some phrases to describe \emph{digestive system diseases}\\
 4.use some sentences  to describe \emph{digestive system diseases}\\
 5.use some phrases to describe \emph{nervous system diseases}\\
 6.use some sentences  to describe \emph{nervous system diseases}\\
 7.use some phrases to describe \emph{cardiovascular disease}\\
 8.use some sentences  to describe \emph{cardiovascular disease}\\
 9.use some phrases to describe \emph{general pathological conditions}}
 \\
 \Xhline{1.2pt}
\end{tabular}
\caption{Prompts to generate candidate concepts on Medical Abstracts dataset.}
\label{candidate_medical_table}
\end{table}

\begin{table}[!htbp]
\begin{tabular}{l}
\Xhline{1.2pt}
 Prompts \\ 
\hline
 \makecell[l]{1.use some  phrases to describe \emph{majority opinion} in legal \\ 
 2.use some  phrases to describe \emph{per curiam} in legal terms\\
 3.use some  phrases to describe \emph{plurality opinion} in legal terms \\
 4.use some  phrases to describe \emph{equally divided} in legal terms\\
 5.use some  phrases to describe \emph{dismissal-moot} in legal terms \\
 6.\makecell[l]{use some  phrases to describe\\ \emph{dismissal-improvidently-granted} in legal terms} \\
 7.\makecell[l]{use some  phrases to describe \emph{dismissal-other} in legal terms}\\
 8.use some  phrases to describe \emph{opinion of the court} \\in legal terms 
 }\\
 \Xhline{1.2pt}
\end{tabular}
\caption{Prompts to generate candidate concepts on Justice dataset.}
\label{candidate_justice_table}
\end{table}

\begin{table}[!htbp]
\begin{tabular}{l}
\Xhline{1.2pt}
 Prompts \\ 
\hline
 \makecell[l]{1.describe the aspects of a \emph{finance positive review} \\ 
 2.what a  \emph{finance positive review} may include\\
 3.describe the aspects of a \emph{finance neutral review} \\
 4.what a  \emph{finance neutral review} may include\\
 5.describe the aspects of a \emph{finance negative review} \\
 6.what a \emph{finance negative review} may include\\
 % 7.\makecell[l]{use some  phrases to describe \emph{dismissal-other} in legal terms}\\
 % 8.use some  phrases to describe \emph{opinion of the court} \\in legal terms \\
 % 7.\makecell[l]{use some  phrases to describe \emph{dismissal-other} in legal terms}\\
 % 8.use some  phrases to describe \emph{opinion of the court} \\in legal terms \\
 % 7.\makecell[l]{use some  phrases to describe \emph{dismissal-other} in legal terms}\\
 % 8.use some  phrases to describe \emph{opinion of the court} \\in legal terms \\
 % 7.\makecell[l]{use some  phrases to describe \emph{dismissal-other} in legal terms}\\
 % 8.use some  phrases to describe \emph{opinion of the court} \\in legal terms \\
 % 7.\makecell[l]{use some  phrases to describe \emph{dismissal-other} in legal terms}
 }\\
 
 \Xhline{1.2pt}
\end{tabular}
\caption{Prompts to generate candidate concepts on Finance Sentiment dataset.}
\label{candidate_justice_table}
\end{table}
% \section{Other Prompts Results}

\begingroup
\setlength{\tabcolsep}{6pt} 
\renewcommand{\arraystretch}{1.0}
\begin{table}[!htbp]
\begin{tabular}{l}
\Xhline{1.2pt}
 Prompts \\ 
\hline
 \makecell[l]{1.[$x$]The feeling of the review is [$p$][$m$].\\ 
 2.[$x$]The feeling of the review is [$p$][$p$][$m$].\\
 3.[$x$]What emotion does the review evoke ? [$p$][$m$]\\
 4.[$x$]What emotion does the review evoke ? [$p$][$p$][$m$]\\
 5.[$x$]describe the sentiment of the text. [$p$][$m$]\\
 6.[$x$]describe the sentiment of the text. [$p$][$p$][$m$]}\\
 \Xhline{1.2pt}
\end{tabular}
\caption{Prompts used in Table\ref{acc_table} on SST-2, IMDB dataset. $m$ is the mask token. $p$ is the prefix token.}
\label{acc_sst2_imdb_table}
\end{table}
% \section{More Explantion Examples}
% \begingroup
% \setlength{\tabcolsep}{1pt} % Default value: 6pt
% % \renewcommand{\arraystretch}{1}
% \begin{table*}[!tb]
% \centering
% % <{\centering}p{1cm}
% \begin{tabular}{llll}
% \Xhline{1.2pt}
% Dataset                 & Class  & Input & Top-3 Concepts \\ \hline
% \multirow{2}{*}[-2ex]{SST-2}  & positive   & \emph{I loved it!}      &  \makecell[l]{1.\emph{Encouraging and Supportive.}\\2.\emph{Happy and Optimistic.}\\3.\emph{Constructive and Helpful.}}               \\ \cline{2-4} 
%                         & negative   &  \emph{extremely confusing.}     &    \makecell[l]{1.\emph{Opposing or contradictory to desired outcomes.}\\2.\emph{Pessimistic outlook or attitude.}\\3.\emph{Discouraging or disheartening experiences.}}                 \\ \Xhline{1.2pt}

% \end{tabular}
% \caption{Explanation examples on SST-2 and AGNews dataset with BERT-Large.}
% \label{more_result_sst2_AGNews_bert}
% \end{table*}
% \endgroup
\section{More Prompt Results}
\label{more_results}
\clearpage

\begingroup
\setlength{\tabcolsep}{1pt} % Default value: 6pt
\renewcommand{\arraystretch}{1.1}
\begin{table*}[!tb]
\centering
\begin{tabular}{llcccccccc}
\toprule[1.2pt]
\multicolumn{2}{c}{\multirow{3}{*}{}} & \multicolumn{4}{c}{BERT-Large}                      & \multicolumn{4}{c}{GPT-2-Medium}                     \\ 
% \cline{3-12} 
\multicolumn{2}{c}{}                  & \multicolumn{2}{c}{$N_q=1$}  & \multicolumn{2}{c}{$N_q=2$}  & \multicolumn{2}{c}{$N_q=1$}  & \multicolumn{2}{c}{$N_q=2$}  \\ 
\cmidrule(lr){3-4}\cmidrule(lr){5-6}\cmidrule(lr){7-8}\cmidrule(lr){9-10}
\multicolumn{2}{c}{}                  & P-tuning & CD  & P-tuning & CD  & P-tuning & CD  & P-tuning & CD  \\ \Xhline{1.2pt}
\multirow{6}{*}[-1.5ex]{ \makecell[l]{What emotion\\ does the\\ review evoke}}    & 4-shot    &  0.527        & 0.518   & 0.568         & 0.560         &0.534   &0.531         & 0.547         &  0.534                    \\ \cline{2-10} 
                          & 8-shot    &  {0.562}        &  0.547  & 0.581         &  0.676        & {0.559}   &  0.577        &   {0.596}       & 0.614              \\ \cline{2-10} 
                          & 16-shot   &   {0.781}       & 0.773   & 0.744         &  0.744        & 0.807   &   {0.813}       &  0.747        & {0.725}                     \\ \cline{2-10} 
                          & 32-shot   &  0.830        & 0.840   &{0.794}          & {0.812}         & 0.843   &   0.843       &   0.825       & {0.819}                     \\ \cline{2-10} 
                          & Full      &   {0.898}       & 0.907   & 0.893         &  0.892        & 0.909   &  {0.911}        &  0.864        & {0.902}          \\ \cline{2-10} 
                          & Avg & \textbf{0.719} & 0.717& \textbf{0.751}& 0.736& \textbf{0.745}& 0.735&0.715&\textbf{0.718} \\ \cline{2-10}
                          & ${\sigma^2}$       &  ${\pm0.003}$        & ${\pm0.001}$   &  ${\pm0.001}$        &  $\pm0.003$        &  $\pm0.003$  & $\pm0.005$         &   $\pm0.0003$       &  $\pm0.002$                \\ \Xhline{1.2pt}
\multirow{6}{*}[-1.5ex]{ \makecell[l]{describe\\ the  sentiment\\ of the text}}     & 4-shot    & 0.545         &0.592    &0.575         & 0.592        &0.549    &0.545          &0.548          &0.542                 \\ \cline{2-10} 
                          & 8-shot    & 0.557         &0.615    &0.576          &0.615         &0.582    &0.593          &0.571        &0.636                    \\ \cline{2-10} 
                          & 16-shot   & 0.718        &0.797    &0.730          &0.797          &0.814    &0.803          &0.771        &0.787          \\ \cline{2-10} 
                          & 32-shot   &0.836         &0.850    & 0.821         &0.850         &0.817    & 0.832        &0.818          &0.843                     \\ \cline{2-10} 
                          & Full      & 0.896         &0.907   &0.895          &0.907          &0.908    &0.909          &0.887        &0.891                     \\ \cline{2-10} 
                          & Avg &0.710  & \textbf{0.752}& 0.719 &\textbf{0.752}&0.734 &0.736 &0.719
                          &\textbf{0.739}\\ \cline{2-10}
                          & $\sigma^2$       &  $\pm0.0007$        & $\pm0.005$   &   $\pm0.003$       & $\pm0.005$         & $\pm0.003$   &    $\pm0.0002$      &  $\pm0.0004$        &  $\pm0.00006$                        \\ \Xhline{1.2pt}
   
\end{tabular}
\caption{Accuracy results of CD with various prompt types on the SST-2 dataset. $N_q$ denotes the number of continuous prompt tokens.  $\sigma^2$ denotes the average variance across five shot settings.}
% \footnote{}
\label{acc_table_sst_addition}
\end{table*}

\begingroup
\setlength{\tabcolsep}{1pt} % Default value: 6pt
\renewcommand{\arraystretch}{1.1}
\begin{table*}[!tb]
\centering
\begin{tabular}{llcccccccc}
\toprule[1.2pt]
\multicolumn{2}{c}{\multirow{3}{*}{}} & \multicolumn{4}{c}{BERT-Large}                      & \multicolumn{4}{c}{GPT-2-Medium}                     \\ 
% \cline{3-12} 
\multicolumn{2}{c}{}                  & \multicolumn{2}{c}{$N_q=1$}  & \multicolumn{2}{c}{$N_q=2$}  & \multicolumn{2}{c}{$N_q=1$}  & \multicolumn{2}{c}{$N_q=2$}  \\ 
\cmidrule(lr){3-4}\cmidrule(lr){5-6}\cmidrule(lr){7-8}\cmidrule(lr){9-10}
\multicolumn{2}{c}{}                  & P-tuning & CD  & P-tuning & CD  & P-tuning & CD  & P-tuning & CD  \\ \Xhline{1.2pt}

\multirow{6}{*}[-1.5ex]{ \makecell[l]{What emotion\\ does the\\ review evoke}}     & 4-shot    &0.551          &0.559    &0.543         &0.543         &0.534    &0.531          & 0.519         &0.525                 \\ \cline{2-10} 
                          & 8-shot    &0.545          &0.546    &0.542          &0.532         &0.559    & 0.577         &0.502        &0.500                    \\ \cline{2-10} 
                          & 16-shot   &0.579         &0.549    & 0.693         &0.663          &0.807    &0.813          &0.501       &0.502           \\ \cline{2-10} 
                          & 32-shot   &0.766         &0.823    &0.818          &0.828         &0.843    & 0.843        &0.700          &0.640                     \\ \cline{2-10} 
                          & Full      &0.908          &0.910   &0.914          &0.914          &0.918    &0.921          &0.902         &0.914                     \\ \cline{2-10} 
                          & Avg &{0.669}  &\textbf{0.677}  &\textbf{0.702}&{0.696}&0.732 &\textbf{0.737} &\textbf{0.624}
                          &0.616\\ \cline{2-10}
                          & $\sigma^2$       &  $\pm0.0009$        & $\pm0.0005$   &   $\pm0.0012$       & $\pm0.0014$         & $\pm0.0003$   &    $\pm0.0004$      &  $\pm0.0009$        &  $\pm0.0012$                        \\ \Xhline{1.2pt}
\multirow{6}{*}[-1.5ex]{ \makecell[l]{describe\\ the  sentiment\\ of the text}}     & 4-shot    &0.549          &0.542    &0.541         &0.592         &0.551    &0.549          & 0.522         & 0.508                \\ \cline{2-10} 
                          & 8-shot    &0.548          &0.528    &0.525          &0.615         &0.517    &0.510          &0.513        &0.501                    \\ \cline{2-10} 
                          & 16-shot   &0.660         &0.620    & 0.660         &0.797          &0.503    &0.501          &0.503        &0.501          \\ \cline{2-10} 
                          & 32-shot   &0.709         &0.698    &0.799          &0.850         &0.714    &0.632         & 0.594         &0.621                     \\ \cline{2-10} 
                          & Full      &0.915          &0.911   &0.919          &0.919          &0.917    &0.927          &0.900         &0.908                     \\ \cline{2-10} 
                          & Avg &\textbf{0.676}  &0.659 &{0.688} &\textbf{0.754}&\textbf{0.640} &0.623 &0.606
                          &0.607\\ \cline{2-10}
                          & $\sigma^2$       &  $\pm0.0021$        & $\pm0.0014$   &   $\pm0.0009$       & $\pm0.0016$         & $\pm0.0004$   &    $\pm0.0002$      &  $\pm0.0007$        &  $\pm0.0003$                        \\ \Xhline{1.2pt}
\end{tabular}
\caption{Accuracy results of CD with various prompt types on the IMDB dataset. $N_q$ denotes the number of continuous prompt tokens.  $\sigma^2$ denotes the average variance across five shot settings.}
% \footnote{}
\label{acc_table_imdb_addition}
\end{table*}

\begingroup
\setlength{\tabcolsep}{1pt} % Default value: 6pt
\renewcommand{\arraystretch}{1.1}
\begin{table*}[!tb]
\centering
\begin{tabular}{llcccccccc}
\toprule[1.2pt]
\multicolumn{2}{c}{\multirow{3}{*}{}} & \multicolumn{4}{c}{BERT-Large}                      & \multicolumn{4}{c}{GPT-2-Medium}                     \\ 
% \cline{3-12} 
\multicolumn{2}{c}{}                  & \multicolumn{2}{c}{$N_q=1$}  & \multicolumn{2}{c}{$N_q=2$}  & \multicolumn{2}{c}{$N_q=1$}  & \multicolumn{2}{c}{$N_q=2$}  \\ 
\cmidrule(lr){3-4}\cmidrule(lr){5-6}\cmidrule(lr){7-8}\cmidrule(lr){9-10}
\multicolumn{2}{c}{}                  & P-tuning & CD  & P-tuning & CD  & P-tuning & CD  & P-tuning & CD  \\ \Xhline{1.2pt}

\multirow{6}{*}[-1.5ex]{ \makecell[l]{The primary focus\\ of this\\ news is}}     & 4-shot    &0.733          & 0.733   & 0.728        & 0.736        &0.746    &0.757          &0.759          &0.758                 \\ \cline{2-10} 
                          & 8-shot    &0.771          & 0.732   &0.779          &0.766         &0.785    &0.776          &0.791        &0.795                    \\ \cline{2-10} 
                          & 16-shot   & 0.800        &0.826    &0.793          &0.791          &0.788    & 0.789         &0.793       & 0.798          \\ \cline{2-10} 
                          & 32-shot   &0.841         &0.844    &0.844          &0.842         &0.825    & 0.824        &0.834          &0.828                     \\ \cline{2-10} 
                          & Full      &0.885          &0.886   &0.874          &0.877          &0.863    &0.867          &0.859         &0.858                     \\ \cline{2-10} 
                          & Avg &{0.806}  &{0.804}  &\textbf{0.822}&{0.802}&{0.801} &{0.802} &0.807
                          &0.807\\ \cline{2-10}
                          & $\sigma^2$       &  $\pm0.0004$        & $\pm0.0004$   &   $\pm0.0001$       & $\pm0.0003$         & $\pm0.0001$   &    $\pm0.0001$      &  $\pm0.0004$        &  $\pm0.0006$                        \\ \Xhline{1.2pt}
\multirow{6}{*}[-1.5ex]{ \makecell[l]{The core theme\\ of this\\ news is}}     & 4-shot    &0.697          &0.777    &0.652         &0.728         &0.778    &0.771          &0.765          &0.778                 \\ \cline{2-10} 
                          & 8-shot    &0.788          &0.752    &0.796          &0.768         &0.795    &0.790          &0.788        & 0.793                   \\ \cline{2-10} 
                          & 16-shot   &0.806         &0.795    &0.810          &0.819          &0.791    &0.801          &0.789       &0.789           \\ \cline{2-10} 
                          & 32-shot   &0.831         &0.831    & 0.829         &0.839         &0.834    &0.840         &0.825          &0.831                     \\ \cline{2-10} 
                          & Full      &0.886          &0.888   &0.876          &0.875          &0.867    &0.865          &0.853         &0.863                     \\ \cline{2-10} 
                          & Avg &{0.801}  &{0.808}  &{0.792}&\textbf{0.805}&0.813 &{0.813} &{0.804}
                          &0.810\\ \cline{2-10}
                          & $\sigma^2$       &  $\pm0.0002$        & $\pm0.0003$   &   $\pm0.0005$       & $\pm0.0004$         & $\pm0.0002$   &    $\pm0.0005$      &  $\pm0.0001$        &  $\pm0.0006$                        \\ \Xhline{1.2pt}
\end{tabular}
\caption{Accuracy results of CD with various prompt types on the AGNews dataset. $N_q$ denotes the number of continuous prompt tokens.  $\sigma^2$ denotes the average variance across five shot settings.}
% \footnote{}
\label{acc_table_imdb_addition}
\end{table*}

\end{document}